\documentclass[sigconf,
                nonacm,
        ]{acmart}
\usepackage{algorithm}
\usepackage{algpseudocode}
\usepackage{amsmath}
\AtBeginDocument{%
  }

\setcopyright{acmlicensed}
\copyrightyear{2024}
\acmYear{2024}
\acmDOI{XXXXXXX.XXXXXXX}

\acmConference[ICAIF '24]{International Conference on AI in Finance}{November 14--16,
  2024}{Brooklyn, NY}
\acmISBN{978-1-4503-XXXX-X/18/06}

\begin{document}

\title[Extracting Structured Insights from Financial News]{Extracting Structured Insights from Financial News:\\An Augmented LLM Driven Approach}

\author{Rian Dolphin}
\orcid{0000-0002-5607-9948}
\affiliation{%
  \country{research@polygon.io}\\
  \institution{Polygon.io}
}

\author{Joe Dursun}
\affiliation{%
  \institution{Polygon.io}
  \city{Atlanta}
  \state{Georgia}
  \country{USA}
}

\author{Jonathan Chow}
\affiliation{%
  \institution{Polygon.io}
  \city{Atlanta}
  \state{Georgia}
  \country{USA}
}

\author{Jarrett Blankenship}
\affiliation{%
  \institution{Polygon.io}
  \city{Atlanta}
  \state{Georgia}
  \country{USA}
}

\author{Katie Adams}
\affiliation{%
  \institution{Polygon.io}
  \city{Atlanta}
  \state{Georgia}
  \country{USA}
}

\author{Quinton Pike}
\affiliation{%
  \institution{Polygon.io}
  \city{Atlanta}
  \state{Georgia}
  \country{USA}
}

\renewcommand{\shortauthors}{Dolphin et al.}

\begin{abstract}
Financial news plays a crucial role in decision-making processes across the financial sector, yet the efficient processing of this information into a structured format remains challenging. This paper presents a novel approach to financial news processing that leverages Large Language Models (LLMs) to overcome limitations that previously prevented the extraction of structured data from unstructured financial news. We introduce a system that extracts relevant company tickers from raw news article content, performs sentiment analysis at the company level, and generates summaries, all without relying on pre-structured data feeds.
Our methodology combines the generative capabilities of LLMs, and recent prompting techniques, with a robust validation framework that uses a tailored string similarity approach. 
Evaluation on a dataset of 5530 financial news articles demonstrates the effectiveness of our approach, with 90\% of articles not missing any tickers compared with current data providers, and 22\% of articles having additional relevant tickers.
In addition to this paper, the methodology has been implemented at scale with the resulting processed data made available through a live API endpoint, which is updated in real-time with the latest news\footnote{
\href{https://polygon.io/docs/stocks/get_v2_reference_news?utm_source=research&utm_campaign=news_sentiment}{https://polygon.io/docs/stocks/get\_v2\_reference\_news}
}. 
To the best of our knowledge, we are the first data provider to offer granular, per-company sentiment analysis from news articles, enhancing the depth of information available to market participants. 
We also release the evaluation dataset of 5530 processed articles as a static file\footnote{
\href{https://www.kaggle.com/datasets/rdolphin/financial-news-with-ticker-level-sentiment}{kaggle.com/datasets/rdolphin/financial-news-with-ticker-level-sentiment}
}, which we hope will facilitate further research leveraging financial news.

\end{abstract}

\ccsdesc[500]{Computing methodologies~Artificial intelligence}
\ccsdesc[500]{Computing methodologies~Natural language processing}
\ccsdesc[500]{Computing methodologies~Information extraction}

\keywords{LLM, Finance, News, Structured Data, NLP}


\maketitle

\section{Introduction}

Financial news plays a crucial role in shaping market sentiment, influencing investment decisions, and driving price movements \cite{tetlock2007giving, garcia2013sentiment}. Timely and accurate analysis of financial news can provide valuable insights for market participants and researchers alike. However, extracting actionable information from the vast amount of unstructured news data remains a significant challenge~\cite{ding2015deep}. The complexities involved in extracting relevant companies, mapping company names to instrument-level identifiers, providing granular sentiment analysis, and navigating legal restrictions on content distribution have limited the scope and quality of news data available to market participants and researchers.

Traditionally, financial data providers have relied on ingesting news through live feeds offered by individual news providers. While functional, this approach has several limitations. Inconsistencies in the structure of these feeds across providers, particularly in the inclusion and format of key information such as keywords or article descriptions, can hinder the efficient extraction of structured data. Moreover, changes in feed structure can disrupt the ingestion process, necessitating frequent adjustments to multiple pipelines to maintain data flow. Perhaps most limiting for use cases that filter news by company is the reliance on pre-tagged instrument identifiers (e.g., tickers) within the live news feeds. This dependency restricts the range of news sources that can be effectively processed, as not all providers include ticker information in their feeds or articles.

These limitations underscore the need for a more robust and flexible system capable of processing raw news content into a consistent format, regardless of its source or pre-existing structure. Recent advancements in natural language processing (NLP), particularly the development of Large Language Models (LLMs), present an opportunity to address these challenges. LLMs have demonstrated remarkable capabilities in understanding and generating human-like text \cite{bubeck2023sparks, anthropic2024claude, kojima2022large}, making them potentially powerful tools for extracting structured information from unstructured news articles.

In this paper, we introduce a novel approach that leverages LLMs to overcome the limitations of traditional financial news processing methods. Our system extracts relevant company tickers directly from raw article content, performs sentiment analysis, and generates summaries, all without relying on pre-structured data feeds. This approach not only broadens the range of usable news sources but also enhances the depth and quality of extracted information.

However, applying LLMs to financial news processing is not without challenges. The dynamic nature of financial markets means that company names, ticker symbols, and other reference data are constantly evolving. Relying solely on the knowledge encoded in LLM weights, which is static after training, can lead to outdated or incorrect information. Moreover, the relationship between company names mentioned in articles and ticker symbols is not always straightforward, particularly for less well-known companies or those with multiple listed entities.

To address these challenges, we introduce a hybrid methodology that combines the generative capabilities of LLMs with a robust validation framework for company-ticker mappings. This framework utilizes a regularly updated database of ticker-company mappings and employs a custom string similarity algorithm to identify and correct potential errors in LLM outputs. This approach ensures the accuracy of extracted information while maintaining the flexibility and power of LLM-based processing.

Our work makes several key contributions to the field of financial news processing. 
\begin{enumerate}
    \item We present a novel hybrid approach that leverages LLMs for unstructured news processing while addressing the specific challenges of company-ticker mapping through a custom validation framework and approximate string matching approach.
    \item We demonstrate strong accuracy and coverage in ticker extraction, with 90\% of articles tagged with at least the same tickers as provided by publishers, and 22\% of articles having additional relevant tickers.
    \item We extract granular, per-company sentiment analysis, along with reasoning, using chain-of-thought prompting \cite{wei2022chain} to improve performance. To the best of our knowledge, we are the first data provider to offer news sentiment at the ticker level.
    \item All resulting data is made available through a live API endpoint, which is updated with the latest news at an hourly cadence\footnote{
    \href{https://polygon.io/docs/stocks/get_v2_reference_news?utm_source=research&utm_campaign=news_sentiment}{https://polygon.io/docs/stocks/get\_v2\_reference\_news}
    }. We also provide a static file with the full dataset of 5530 processed articles that can be easily used by researchers\footnote{
    \href{https://www.kaggle.com/datasets/rdolphin/financial-news-with-ticker-level-sentiment}{kaggle.com/datasets/rdolphin/financial-news-with-ticker-level-sentiment}
    }.
\end{enumerate}

The data and insights made available through our LLM-powered news processing pipeline have the potential to benefit both the research community and industry practitioners. By implementing the system at scale, we have created a comprehensive dataset of news articles with relevant tickers and sentiment tags. By making this data accessible and continually updated, we aim to spur further advancements in the application of AI to financial markets.

The remainder of this paper is organized as follows. Section \ref{sec:related_work} reviews related work in financial news processing and the application of LLMs in finance. Section \ref{sec:methodology} details our methodology for processing financial news articles using LLMs, including data collection, initial LLM processing, ticker validation, and data enrichment. Section \ref{sec:results} presents our experimental results, demonstrating the effectiveness of our approach in identifying relevant tickers and providing comprehensive tagging. Finally, Section \ref{sec:discussion} discusses the implications of our work, potential future directions, and concludes the paper.

\section{Related Work}\label{sec:related_work}

The analysis of financial news and its impact on markets has been a subject of extensive research~\cite{tetlock2010does, liu2018stock}. Seminal work by \citet{tetlock2007giving} demonstrated the relationship between media pessimism and downward pressure on market prices, highlighting the importance of news sentiment in predicting market movements. Building on this foundation, \citet{loughran2011liability} developed a comprehensive financial sentiment dictionary, which has become a standard tool in the field for analyzing the tone of financial texts.

Early approaches to financial news processing relied heavily on rule-based systems and hand-crafted features. These methods, while interpretable, often struggled with the complexity and ambiguity of natural language. As machine learning techniques advanced, researchers began applying supervised learning algorithms to news classification and sentiment analysis tasks. \citet{nassirtoussi2014text} provided a comprehensive survey of text mining techniques for market prediction, highlighting the transition from traditional statistical methods to more sophisticated machine learning approaches.

The application of NLP techniques in finance has seen significant growth in recent years. Word embedding models, such as Word2Vec~\cite{mikolov2013efficient} and GloVe~\cite{pennington2014glove}, have been adapted for financial text analysis, allowing for more nuanced representations of financial terminology. 
\citet{du2020stock} proposed a stock embedding model such that the embedding of related text exhibits high similarity with the embedding of the stock. Similarly, \citet{dolphin2022machine} learn stock embeddings based on co-occurrence of tickers mentioned in news articles using a similar model architecture to Word2Vec. Deep learning models have also made substantial inroads in financial news processing~\cite{liu2018stock,vargas2017deep,ma2019news2vec}. Ding et al. ~\cite{ding2015deep} proposed a deep learning method for event-driven stock prediction, demonstrating the potential of neural networks in capturing complex relationships between news events and stock price movements.

The advent of LLMs has opened the door to numerous potential applications within the financial domain~\cite{li2023large}. Early LLMs fine-tuned on financial text, like FinBERT~\cite{araci2019finbert}, had previously led the way in financial sentiment analysis tasks compared to base models. However, recent large foundation models like GPT4~\cite{achiam2023gpt} and Claude~\cite{anthropic2024claude} consistently outperform domain specific fine-tuned models at niche tasks~\cite{bubeck2023sparks}. Subsequently, researchers and industry practitioners have leveraged LLMs for a range of financial applications from sentiment analysis to credit risk reporting~\cite{xing2024designing,zhang2023enhancing,zeng2023flowmind,teixeira2023enhancing,loukas2023making}. 

Accurate extraction of company names mentioned in news articles and their subsequent mapping to a widely accepted instrument level identifier is crucial for efficient financial news processing. Recent work by \citet{zhang2023finbert} on FinBERT-MRC has shown promise in named entity recognition (NER) for financial texts, potentially improving the accuracy of company name extraction from news articles. However, again large foundation models have been shown to outperform domain specific models at NER tasks~\cite{bubeck2023sparks}. Recent work has explored various techniques, like retrieval-augmented generation technique as well as metadata, to reduce hallucination in extracting information from
financial reports when using LLMs~\cite{sarmah2023towards}. Furthermore, assuming a robust NER pipeline for company name extraction from unstructured text, the task of linking these extracted names to the correct ticker symbols remains an open challenge.

Our work builds upon these learnings, leveraging the power of LLMs for comprehensive financial news processing while addressing the specific challenges of company-ticker mapping through a novel hybrid approach. By combining the generative capabilities of LLMs with a robust validation framework, we aim to overcome the limitations of existing methods and provide a flexible and accurate system for extracting valuable structured data from unstructured financial news. This approach not only enhances the depth and quality of extracted information but also broadens the range of usable news sources, potentially spurring further advancements in the application of AI to financial markets.

\section{Methodology}\label{sec:methodology}

Our methodology for processing financial news articles using LLMs consists of several key steps. 
This section provides a detailed description of each step in the pipeline.


\subsection{Data Collection}

The first step in our pipeline is to collect financial news articles from a broad range of news providers. We achieve this by fetching a live aggregate news feed from Google News for a select list of providers and keywords. For example, we can use query parameters in the URL to obtain articles related to the technology sector from Yahoo Finance published within the last hour.

\begin{figure}
    \centering
    \includegraphics[width=\linewidth]{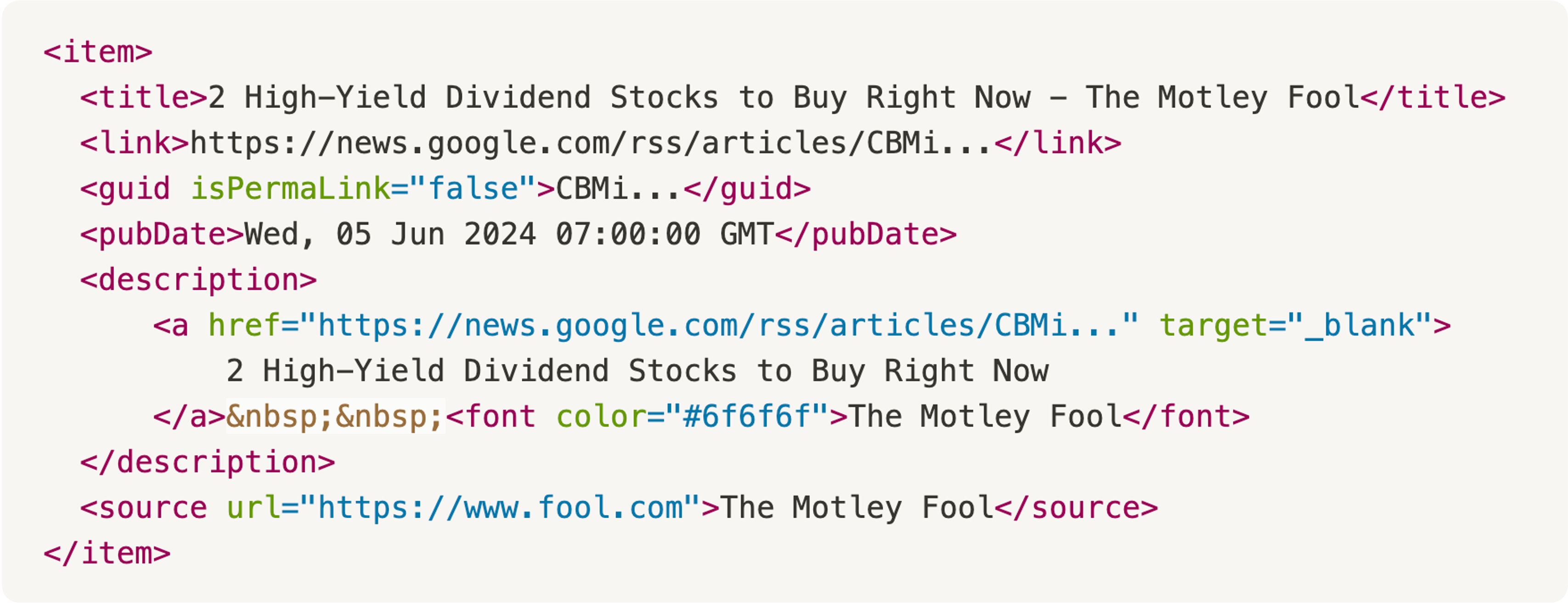}
    \caption{Example article from the Google News live feed.}
    \label{fig:google-rss}
\end{figure}

The live news feed is parsed to extract individual news items, each containing a title, link, publication date, and other metadata. An example of an extracted item is shown in Figure \ref{fig:google-rss}. Each extracted article has a link on the \texttt{news.google.com} domain. By following redirects, we can obtain the final article URL on the news providers site, which is used to fetch the full article content for further processing.

\subsection{Initial LLM Processing}

\begin{figure*}
    \centering
    \includegraphics[width=\linewidth]{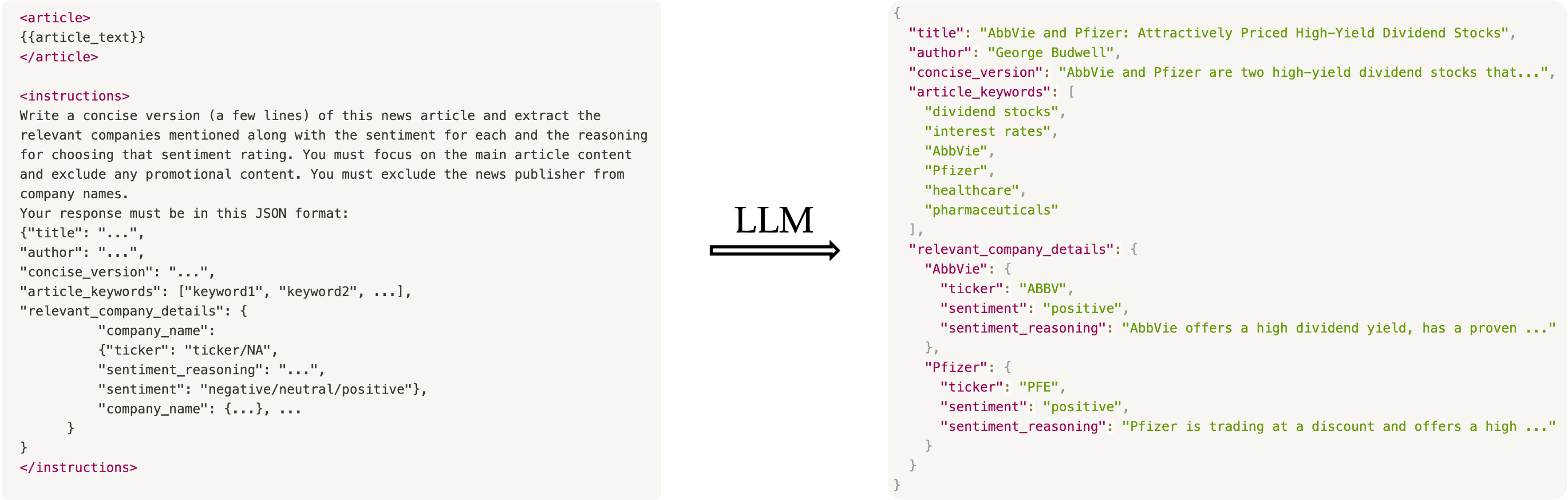}
    \caption{Initial prompt and output example showing the extraction of structured data from an article.}
    \label{fig:prompt1}
\end{figure*}

With the article URL and content, we make an initial call to the LLM to extract key details from the article. The LLM is prompted to provide a structured output containing the title, summary, keywords, relevant companies and associated sentiment details. An example of the prompt and corresponding output is shown in Figure \ref{fig:prompt1}.

When engineering the prompt, we consulted the literature and applied a number of proposed techniques to improve performance. Following \citet{anthropic2024claude}, we distinctly separate the article text and instructions using XML syntax for ease of analysis by the LLM. For the sentiment portion of the prompt, we purposely indicate that the sentiment reasoning should precede the actual sentiment classification. Asking the model for the intermediate reasoning step follows the idea of chain-of-thought prompting outlined in \citet{wei2022chain}, which has been shown to improve model performance.
In addition to using the prompt to promote structured output, we also use partially-filled responses that begin with the ``\texttt{\{}'' character. This results in the generated output skipping filler text and improving instruction following with regard to producing JSON output.

In Figure \ref{fig:prompt1}, we purposely show an example where the LLM incorrectly hallucinates the ticker symbol for Pfizer as \texttt{PFI} instead of \texttt{PFE}. This is chosen to highlight the need for additional validation steps to ensure the accuracy of the extracted information. We will use this article as an example in Section \ref{sec:ticker_validation} when detailing our company-ticker mapping validation pipeline. 

\subsection{Ticker Validation}\label{sec:ticker_validation}
Validating the ticker symbols generated by the LLM is an essential part of the pipeline. An LLM can extract company names that are written in the article with minimal risk of hallucination, since the tokens simply need to be repeated rather than reasoned about. In other words, it is an NER task rather than a reasoning one. However, the hallucination risk of generating tickers (which are often not mentioned in the article) is much higher since we are now relying on the weights of the LLM to have encoded the mapping from company name to ticker.

To validate the generated ticker symbols, we use a dataset of company name to ticker mappings, which is also available via API endpoint\footnote{
\href{https://polygon.io/docs/stocks/get_v3_reference_tickers__ticker?utm_source=research&utm_campaign=news_sentiment}{https://polygon.io/docs/stocks/get\_v3\_reference\_tickers\_\_ticker}
}.
For example, querying the API with the ticker \texttt{ABBV} returns:
\begin{verbatim}
{'ticker': 'ABBV',
  'name': 'ABBVIE INC.',
  'market': 'stocks',
  'locale': 'us',
  ...}
\end{verbatim}
Similarly, querying the API with the incorrect ticker \texttt{PFI} returns:
\begin{verbatim}
{'ticker': 'PFI',
  'name': 'Invesco Dorsey Wright Financial Momentum ETF',
  'market': 'stocks',
  'locale': 'us',
   ...}
\end{verbatim}

This data is leveraged through another LLM call where we ``ask'' the LLM to verify if the company name generated by the original response matches the true company name of the generated ticker. In the example shown in Figure \ref{fig:prompt1}, we are essentially asking \textit{``Does AbbVie match ABBVIE INC.?''} and \textit{``Does Pfizer match Invesco Dorsey Wright Financial Momentum ETF?''}. In this case, the LLM's structured output will indicate that \texttt{ABBV} is a match for AbbVie, but \texttt{PFI} does not match Pfizer.

\subsection{Algorithmic Company-Ticker Mapping}

\begin{algorithm}[t]
\caption{Company-Ticker Mapping Algorithm}
\label{alg:company_ticker_mapping}
\begin{algorithmic}[1]
\Require Query company name $Q$, Dictionary of company names to tickers $N2T$
\Ensure Best matching company name and ticker
\State $Q_{clean} \gets \texttt{RemoveJunkWords}(Q.\text{toLowerCase}())$
\State $data \gets []$
\For{each $(name, ticker)$ in $N2T$}
    \State $name_{lower} \gets name.\text{toLowerCase}()$
    \State $name_{clean} \gets \texttt{RemoveJunkWords}(name_{lower})$
    \State $lev_{clean} \gets \texttt{LevenshteinDistance}(Q_{clean}, name_{clean})$
    \State $lcs_{clean} \gets \texttt{LongestCommonSubstring}(Q_{clean}, name_{clean})$
    \State $common_{words} \gets \texttt{CountCommonWords}(Q_{clean}, name_{clean})$
    \State $data.\text{append}((name, ticker, lev_{clean}, lcs_{clean}, common_{words}))$
\EndFor
\State $candidates \gets \texttt{FilterByMaxLCS}(data)$
\State $best\_match \gets \texttt{SortCandidates}(candidates)$
\State \Return $best\_match$
\end{algorithmic}
\end{algorithm}

For company names that do not match the ticker returned by the LLM, we now want to find the matching tickers. We use an updated database of over 42,000 company name and ticker mapping pairs. The goal is to match the company name mentioned in the news article, and extracted by the LLM, to one of the company names in the database for which we know the corresponding ticker. This can be seen as an approximate string matching problem, and in this section we outline a tailored algorithmic approach to tackle it.
The proposed approach combines multiple string similarity metrics and heuristics to achieve robust company-ticker mappings. The process is outlined in Algorithm \ref{alg:company_ticker_mapping} and the key details are discussed further below.

\begin{enumerate}
    \item \textit{Junk Word Removal:} We preprocess company names by removing common suffixes (e.g., ``Corp.'', ``Inc.'', ``Class A'') and converting to lowercase. This step helps normalize company names and reduce noise in the matching process.
    
    \item \textit{Multiple Similarity Metrics:} We calculate several string similarity metrics for each potential match:
    \begin{itemize}
        \item Levenshtein distance~\cite{levenshtein1966binary} is applied on the cleaned company names. It calculates the minimum number of single-character edits (insertions, deletions, or substitutions) required to change one string into another.
        \item Length of the longest common substring (LCS) on the cleaned company names.
        \item Number of common words between the cleaned query and potential match.
    \end{itemize}
    
    \item \textit{Multi-criteria Sorting:} We first filter candidates based on the maximum LCS length. Among the candidates with the maximum LCS, we further sort based on two criteria: the number of common words (in descending order) and the Levenshtein distance of the cleaned names (in ascending order). This multi-criteria approach helps to balance between partial word matches and overall string similarity.
\end{enumerate}

The algorithm returns the best matching ticker for each company name. However, some mappings are infeasible to solve with this approximate string matching solution; for example, articles mentioning Google refer to the company Alphabet. To mitigate potential errors, especially for these non-obvious mappings, we implement additional safeguards. For example, the distribution of missing tickers by company is monitored to identify problematic mappings, particularly for companies receiving high request volume from customers. If problem tickers are identified we can manually add helpful mappings into the database (i.e. $\texttt{Alphabet}\mapsto\texttt{GOOG}$) to prevent future exclusion.

Finally, after the algorithm returns the most likely ticker for each initially mismatched company name, we make another validation call to the LLM to protect against false positive mappings. If the match is confirmed, we use the corresponding ticker; otherwise, the company name is discarded to maintain data integrity. In the running example from Figure \ref{fig:prompt1}, this process correctly maps Pfizer to the ticker \texttt{PFE}.

We note here that choosing the correct string similarity algorithm to rank prospective tickers is a challenging task. Even after much experimentation around string similarity metrics and various weighted combinations, there are some examples that remain infeasible for a generic fuzzy string matching solution. A prime example of this is news articles mentioning Google, since the corresponding parent company name is Alphabet. As a result, if the LLM were to generate an incorrect ticker, the string similarity algorithm would also fail to match Google to its ticker(s).

We took a number of steps to mitigate the problem of failed mappings for specific companies. Firstly, we examined the distribution of missing tickers by company so that any company whose mapping failed consistently could be identified. Secondly, we did manual checks for tickers that receive high search volume from our users. We also consider that more popular tickers are also those where the LLM is more likely to have encoded the correct company-ticker mappings. The experimental results in Section \ref{sec:results} also provide confidence that these steps have helped mitigate the issue.

\subsection{Data Enrichment and Storage}\label{sec:data_enrichment}

\begin{figure*}[t]
    \centering
    \includegraphics[width=0.65\textwidth]{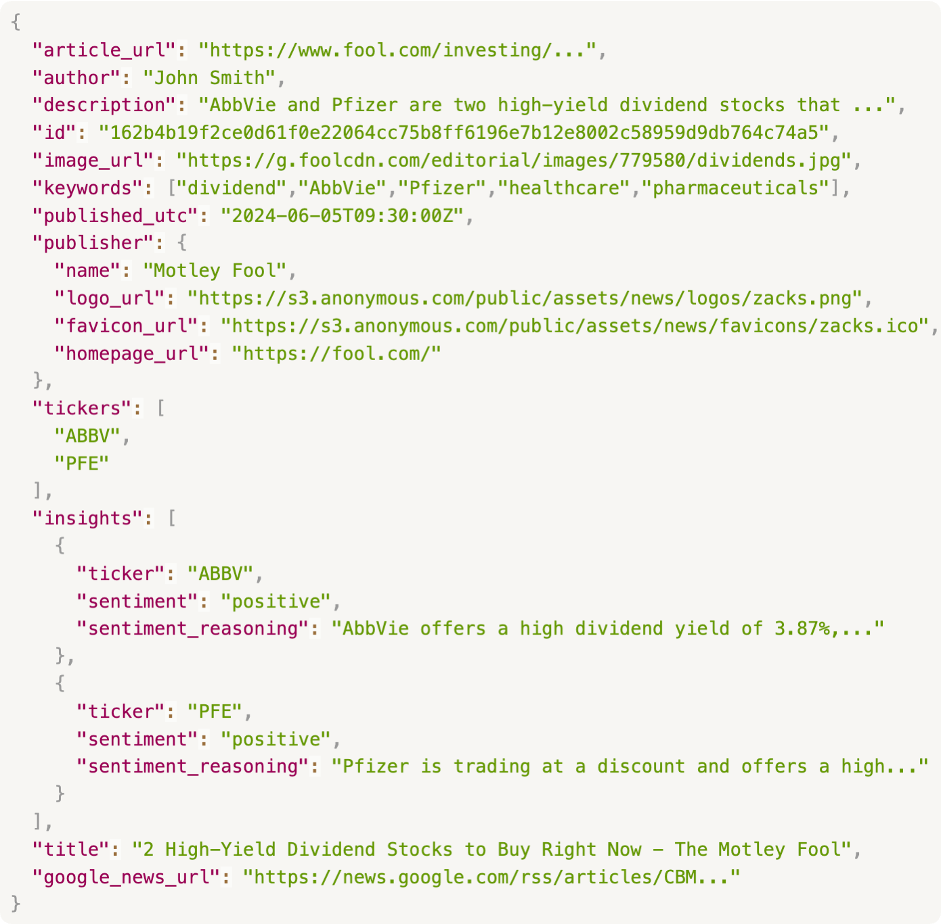}
    \caption{Example of the final output accessible to users via API.}
    \label{fig:final_output}
\end{figure*}

With the validated tickers and sentiment information, we enrich the generated article data with additional information and data fields. Firstly, as we are focusing on tickers as the identifier of choice, and tickers are not unique at the instrument level, we include a post-processing step that adds any tickers which are associated with the same CIK code as the identified relevant tickers. This means that if the system has identified \texttt{GOOG} (the Class A shares of Alphabet) as a relevant ticker, we will automatically include \texttt{GOOGL} (the Class C shares) since they both correspond to the same CIK identifier. 

In addition to this post processing step, we also include new data fields such as the article URL, publication date, image URL and publisher details. The final output for each article is shown in Figure \ref{fig:final_output}. This enriched data is then stored in a database and made available through an API endpoint for downstream applications and further research.

\section{Experimental Results}\label{sec:results}

\begin{figure}
    \centering
    \includegraphics[width=\linewidth]{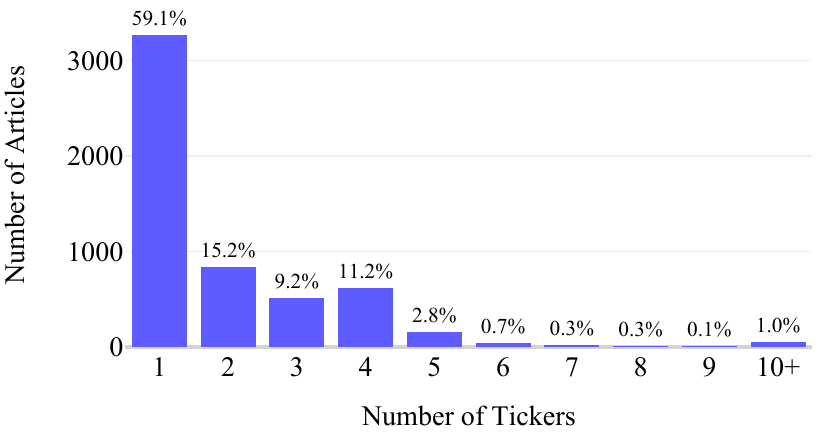}
    \caption{Distribution of number of tickers per article.}
    \label{fig:article-ticker-distribution}
\end{figure}

To evaluate the proposed approach for tagging articles with relevant tickers, we compared against the tickers provided by the news publisher of each article. In total, we selected a random sample of articles from 2023 that spanned six news providers. The choice of providers was limited to those who offered relevant tickers with each articles. We also limited our analysis to common stock and American depository receipts (ADRs), which excluded tickers relating to ETFs and cryptocurrencies. The final dataset contains 5530 news articles and is made available for download\footnote{
\href{https://www.kaggle.com/datasets/rdolphin/financial-news-with-ticker-level-sentiment}{kaggle.com/datasets/rdolphin/financial-news-with-ticker-level-sentiment}
}.
Figure \ref{fig:article-ticker-distribution} shows the distribution of how many tickers are present in each article. We can see that the majority of articles (59\%) are tagged with just one relevant ticker, and 94\% of articles contain 4 or less relevant tickers.

\begin{figure}
    \centering
    \includegraphics[width=\linewidth]{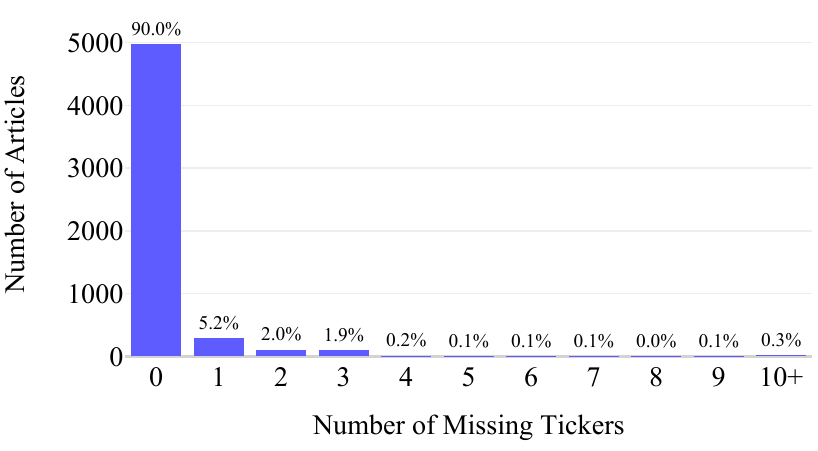}
    \caption{Distribution of the number of missing tickers in articles compared with news provider labelling.}
    \label{fig:missing-histogram}
\end{figure}

To assess the performance of our system, we compared the tickers generated by our approach to those provided by the news publishers. Figure \ref{fig:missing-histogram} illustrates the distribution of missing tickers, i.e., tickers that were included by the news provider but not by our system. Notably, 90\% of articles had no missing tickers, indicating that our system successfully identified all relevant tickers for the vast majority of articles. The percentage of articles with missing tickers decreases rapidly as the number of missing tickers increases, with only 5\% having one missing ticker and 2\% having two.

Manual inspection of the missing tickers revealed that some of these ``misses'' were actually desirable. For example, in an article about Axsome Therapeutics, Jazz Pharmaceuticals was mentioned once in the context of Axsome acquiring a drug from them. Our system did not tag this article with the ticker \texttt{JAZZ}, as the article's primary focus was not on Jazz Pharmaceuticals. Similarly, some providers include tickers for companies mentioned in unrelated sections of the article, such as the section \textit{``Zacks Rank \& Other Key Picks''} often included in articles from \texttt{zacks.com}, which are not directly relevant to the main content. Not including these tickers is preferable and aligns with the goal of providing accurate and relevant ticker tags to users.

Legitimate missed tickers were rare and primarily occurred in articles discussing a large number of less well-known companies, such as ``today's big movers'' style articles. This aligns with the limitations discussed in the methodology, where the LLM-based ticker mapping and string matching approach may fail for less prominent companies.

\begin{figure}
    \centering
    \includegraphics[width=\linewidth]{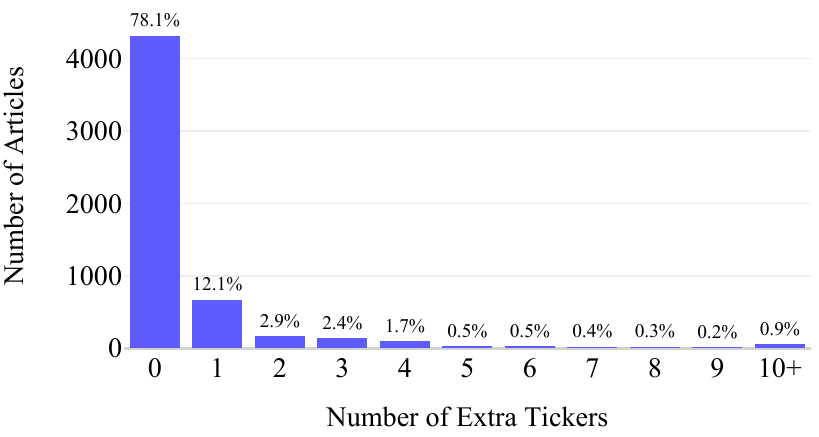}
    \caption{Distribution of the number of additional tickers in articles compared with provider labelling.}
    \label{fig:extra-histogram}
\end{figure}

Figure \ref{fig:extra-histogram} shows the distribution of additional tickers, i.e., tickers that were included by our system but not by the news provider. 78\% of articles had no additional tickers, while 12\% had one extra ticker. The vast majority of these additional tickers result from the post-processing step described in Section \ref{sec:data_enrichment}, which adds all tickers with the same CIK code as any ticker identified by the AI pipeline. This ensures that all share classes and related securities, such as stock warrants, are comprehensively tagged. For example, articles that may only be tagged with \texttt{GOOG} by current news providers will be tagged with both \texttt{GOOG} and \texttt{GOOGL} by our system.

These results demonstrate that our proposed approach, leveraging LLMs and a robust validation framework, effectively identifies relevant tickers for financial news articles. The system achieves high accuracy in matching the tickers provided by news publishers while also providing more comprehensive tagging by including related securities. The occasional ``missing'' tickers are often due to differences in tagging philosophy rather than system failures, and our approach helps to ensure that the most relevant and important tickers are consistently identified.

\section{Discussion and Conclusion}\label{sec:discussion}

The approach presented in this paper overcomes several key challenges that have previously limited the ability of financial data providers to supply high-quality news data to customers. By leveraging the power of LLMs to extract structured data directly from raw news content, we have removed the dependence on pre-tagged ticker information from news providers. This has allowed us to dramatically expand the number of supported news sources by over 400\%, giving users of the data access to a wider range of high-quality news.

A key innovation of our work is providing sentiment analysis at the ticker level for news articles. To the best of our knowledge, we are the first data provider to offer this granular sentiment data from financial news articles, representing a significant value-add for customers. By applying chain-of-thought prompting to improve the LLM's sentiment analysis capabilities and carefully validating the ticker mappings, we are able to generate reliable, ticker-specific sentiment scores that can power new products and trading strategies.

In addition to ticker-level sentiment, our approach generates concise article summaries that can be provided directly to customers. This capability helps address the challenge of legal restrictions that prevent the distribution of full article content from certain news publishers. By creating original summaries that capture the key points of each article, we can supply users with informative article synopses without running into content distribution limitations.

The data and insights made available through our novel news processing pipeline have the potential to benefit the research community and spur further advancements in the application of AI to financial markets. There is already a rich body of literature leveraging financial news data, from exploring news-based sentiment signals to mining company relationships from news co-occurrence~\cite{tetlock2007giving, garcia2013sentiment, ding2015deep, dolphin2022machine}. The approach presented in this paper has enabled us to create a comprehensive dataset of news articles with relevant tickers and sentiment tags. By making the processed data available and continually updated, we hope to promote further research in the application of AI to financial news.

Looking ahead, we see several promising directions for future work. One avenue is to explore the integration of our news processing pipeline with other alternative data sources, such as corporate filings or earnings call transcripts. By combining insights from multiple unstructured data types, we can improve the scope and quality of data we provide.

In conclusion, our LLM-powered approach to financial news processing represents a major step forward in providing customers with comprehensive, high-quality news data in a structured format. By overcoming key challenges around ticker mapping, sentiment analysis, and content distribution, we have created a powerful new data source for both market participants and researchers.


\bibliographystyle{ACM-Reference-Format}
\bibliography{sample-base}

\end{document}